\documentclass[letterpaper]{article} 
\usepackage{aaai24}  
\usepackage{times}  
\usepackage{helvet}  
\usepackage{courier}  
\usepackage[hyphens]{url}  
\usepackage{graphicx} 
\usepackage{natbib}  
\usepackage{caption} 
\usepackage{booktabs}
\usepackage{bbding}
\usepackage{color}
\usepackage{amssymb}
\usepackage{amsmath}
\usepackage{multirow}
\usepackage[dvipsnames]{xcolor}
\definecolor{DarkGreen}{RGB}{1,100,32} 
\usepackage{algorithm}
\usepackage{algorithmic}
\newtheorem{definition}{Definition}

\usepackage{newfloat}
\usepackage{listings}
\DeclareCaptionStyle{ruled}{labelfont=normalfont,labelsep=colon,strut=off} 
\floatstyle{ruled}
\newfloat{listing}{tb}{lst}{}
\floatname{listing}{Listing}
\title{Towards Effective and General Graph Unlearning via Mutual Evolution}
\author{
    Xunkai Li\equalcontrib\textsuperscript{\rm 1},
    Yulin Zhao\equalcontrib\textsuperscript{\rm 3},
    Zhengyu Wu\textsuperscript{\rm 1},\\
    Wentao Zhang\textsuperscript{\rm 4, 5},
    Rong-Hua Li\textsuperscript{\rm 1, 2},
    Guoren Wang\textsuperscript{\rm 1}
}
\affiliations{
    \textsuperscript{\rm 1}Beijing Institute of Technology, Beijing, China\\
    \textsuperscript{\rm 2}Shenzhen Institute of Technology, Shenzhen, China\\
    \textsuperscript{\rm 3}Shandong University, Shandong, China\\
    \textsuperscript{\rm 4}Peking University, Beijing, China\\
    \textsuperscript{\rm 5}National Engineering Labratory for Big Data Analytics and Applications, Beijing, China\\
    \{cs.xunkai.li, yulinzhao233\}@gmail.com, Jeremywzy96@outlook.com,\\wentao.zhang@pku.edu.cn, lironghuabit@126.com, wanggrbit@gmail.com

}

\usepackage{bibentry}

\begin{document}

\maketitle

\begin{abstract}
    With the rapid advancement of AI applications, the growing needs for data privacy and model robustness have highlighted the importance of machine unlearning, especially in thriving graph-based scenarios.
    However, most existing graph unlearning strategies primarily rely on well-designed architectures or manual process, rendering them less user-friendly and posing challenges in terms of deployment efficiency.
    Furthermore, striking a balance between unlearning performance and framework generalization is also a pivotal concern.
    To address the above issues, we propose \underline{\textbf{M}}utual \underline{\textbf{E}}volution \underline{\textbf{G}}raph \underline{\textbf{U}}nlearning (MEGU), a new mutual evolution paradigm that simultaneously evolves the predictive and unlearning capacities of graph unlearning. 
    By incorporating aforementioned two components, MEGU ensures complementary optimization in a unified training framework that aligns with the prediction and unlearning requirements.
    Extensive experiments on 9 graph benchmark datasets demonstrate the superior performance of MEGU in addressing unlearning requirements at the feature, node, and edge levels. 
    Specifically, MEGU achieves average performance improvements of 2.7\%, 2.5\%, and 3.2\% across these three levels of unlearning tasks when compared to state-of-the-art baselines. 
    Furthermore, MEGU exhibits satisfactory training efficiency, reducing time and space overhead by an average of 159.8x and 9.6x, respectively, in comparison to retraining GNN from scratch.

\end{abstract}

\section{Introduction}
\label{sec: introduction}
    Recently, graphs have been a trending AI topic. 
    To enable graph learning with human-like intelligence, graph neural networks (GNNs) have achieved state-of-the-art performance in node-~\cite{chen2020gcnii,gamlp}, link-~\cite{cai2021link_prediction2,tan2023link_prediction4}, and graph-level~\cite{xu2018gin,yang2022graph_classification3} scenarios.

    As most academic works center on training GNN under experimental settings, its real-world implementation often requires extra modifications to meet practical demands, such as the deletion of graph elements.
    It is critical in practicing data-driven AI applications, where the presence of irrelevant, inaccurate, or privacy-sensitive data elements can significantly impact the predictive performance of trained GNNs.
    Two motivations behind the real-world AI deployment of data deletion can be further illustrated as follows:
    (i) \textbf{Data privacy}: Deletion of elements takes into account the "right to be forgotten" in machine learning, enabling users to request the removal of sensitive elements used for training. 
    As a result, this changes node presence and helps protect data privacy.
    (ii) \textbf{Model robustness}: 
    The presence of industry-related noise and fluctuation compromises data quality. 
    By employing data deletion, the impact of such noise on contaminating node attributes and edge presence can be mitigated, leading to enhanced model robustness.

    To achieve data deletion, machine unlearning (MU) is introduced, aiming to enable trained models to forget the influence of unlearning entities (deleted elements).
    In general, the MU strategy contains two crucial modules for practical demands:
    (i) \textbf{Predictive module}: It maintains predictive performance for non-unlearning entities;
    (ii) \textbf{Unlearning module}: It removes the influence of unlearning entities.
    Given the distinctive graph-based challenges in real-world deployments, addressing fundamental tasks of graph unlearning (GU) involves designing strategies for feature, node, and edge-level operations. 
    Compared to MU in computer vision, GU poses unique challenges since the extensive entity interactions by GNN training (i.e., message-passing).
    A naive approach is to retrain the model from scratch but it suffers from the high costs of frequent unlearning requests.
    
    Recently, some approximate-based GU methods are proposed.
    GIF~\cite{wu2023gif} establishes the graph influence function to capture the relationship between data variations and model weights, and certified GU approaches~\cite{chien2022cgu, chien2022sgc_unlearning} propose a theoretical framework for approximate unlearning in linear GNNs.
    These methods mainly focus on the unlearning module but overlook the predictive module.
    As a result, although these methods offer high flexibility, related research~\cite{mitchell2021mend} highlights potential compromises in their practical performance due to limited consideration for non-unlearning entities.
    Meanwhile, seeking a balanced trade-off between generalization boundaries and performance remains challenging in real world deployment.
    Other GU approaches~\cite{chen2022graph_eraser,cong2023projector,cheng2023gnndelete,wang2023guide} introduce learnable mechanisms to adjust the original model or output for non-unlearning entities while eliminating the impact of unlearning entities. 
    However, their predictive and unlearning capabilities often rely on well-designed architectures and handcrafted mechanisms, leaving room for improvement.

\begin{table}[t]
\resizebox{\linewidth}{15mm}{
\setlength{\tabcolsep}{1.2mm}{
\begin{tabular}{cccccc}
\hline
Methods     & Types  & \begin{tabular}[c]{@{}c@{}}Model \\ Agnostic?\end{tabular} & \begin{tabular}[c]{@{}c@{}}Preserve\\ Performance?\end{tabular} & \begin{tabular}[c]{@{}c@{}}Continue\\ Training?\end{tabular}  & \begin{tabular}[c]{@{}c@{}}Deploy\\ Efficiency?\end{tabular} \\ \hline
GIF (Wu et al. 2023)         & Appro. & \textcolor{black}{\Checkmark}                                    & \textcolor{black}{\XSolidBrush}                                      & \textcolor{black}{\Checkmark}                                 & \textcolor{black}{\Checkmark}                                       \\
CGU (Pan et al. 2022)         & Appro. & \textcolor{black}{\XSolidBrush}                                  & \textcolor{black}{\XSolidBrush}                                      & \textcolor{black}{\Checkmark}                                 & \textcolor{black}{\XSolidBrush}                                        \\
GUIDE (Wang et al. 2023)        & Learn. & \textcolor{black}{\Checkmark}                                   & \textcolor{black}{\Checkmark}                                      & \textcolor{black}{\Checkmark}                                 & \textcolor{black}{\XSolidBrush}                                       \\
Projector (Cong et al. 2023)   & Learn. & \textcolor{black}{\XSolidBrush}                                  & \textcolor{black}{\Checkmark}                                      & \textcolor{black}{\XSolidBrush}                                 & \textcolor{black}{\Checkmark}                                       \\
Delete (Cheng et al. 2023)   & Learn. & \textcolor{black}{\Checkmark}                                   & \textcolor{black}{\Checkmark}                                      & \textcolor{black}{\XSolidBrush}                                  & \textcolor{black}{\Checkmark}                                        \\
Eraser (Chen et al. 2022) & Learn. & \textcolor{black}{\Checkmark}                                   & \textcolor{black}{\Checkmark}                                      & \textcolor{black}{\Checkmark}                                 & \textcolor{black}{\XSolidBrush}                                       \\ \hline
MEGU (This Paper)        & Learn. & \textcolor{black}{\Checkmark}                                   & \textcolor{black}{\Checkmark}                                      & \textcolor{black}{\Checkmark}                                 & \textcolor{black}{\Checkmark}                                       \\ \hline
\end{tabular}
}}
\caption{A summary of recent GU studies. 
}
\label{tab: gu_methods}
\end{table}

    Building upon this, we review recent GU methods in Table~\ref{tab: gu_methods} and suggest that a successful GU method should be capable of both handling unlearning requests at any time and being applicable to any backbone model (Model Agnostic). 
    Hence, it should not only generate predictions that prioritize the performance of non-unlearning entities (Preserve Performance) but also possess the ability to adjust the trained model and continue training (Continue Training). 
    Notably, the focus should be on designing these processes with a priority on mitigating the impact of unlearning entities.
    Furthermore, considering the real-world deployment requirements, they should demonstrate high efficiency in the both training and inference process (Deploy Efficiency).
    
    \textbf{Our contributions.}
    (1) \textit{\underline{New Perspective}}. 
    In this paper, we first emphasize the constraints of current GU strategies from a new perspective involving two distinct modules.
    Then, we provide a comprehensive review in Table~\ref{tab: gu_methods} to clarify the design target of GU.
    (2) \textit{\underline{New Method}}. 
    Building upon this, we propose \underline{\textbf{M}}utual \underline{\textbf{E}}volution \underline{\textbf{G}}raph \underline{\textbf{U}}nlearning (MEGU), which comprises original model-based predictive module and linear unlearning module to adjust the original model and generate predictions for non-unlearning entities, respectively.
    From the mutual evolution perspective, the effectiveness of the predictive module in eliminating the influence of unlearning entities relies on the forgetting capability of the unlearning module, and the reasoning capability of the predictive module is essential for the unlearning module to generate reliable predictions.
    (3) \textit{\underline{SOTA Performance}}. 
    Extensive experiments on 9 benchmark datasets demonstrate that MEGU achieves not only state-of-the-art performance but also high training efficiency and scalability.
    Especially, MEGU outperforms GNNDelete~\cite{cheng2023gnndelete} by a margin of 2.8\%-6.4\% in terms of predictive accuracy, while achieving up to 4.5×-7.2× training speedups, respectively.

\section{Preliminaries}
\subsection{Problem Formalization}
    In this work, we focus on the semi-supervised node classification task based on the topology of labeled set $\mathcal{V}_L$ and unlabeled set $\mathcal{V}_U$, and the nodes in $\mathcal{V}_U$ are predicted with the supervised by $\mathcal{V}_L$.
    Consider a graph $\mathcal{G} = (\mathcal{V}, \mathcal{E}, \mathcal{X})$ with $|\mathcal{V}|=n$ nodes, $|\mathcal{E}|=m$ edges, and $\mathcal{X}=\mathbf{X}$.
    The feature matrix is $\mathbf{X} = \{x_1,\dots,x_n\}$ in which $x_v\in\mathbb{R}^{f}$ represents the feature vector of node $v$, and $f$ represents the dimension of the node attributes, the adjacency matrix (including self-loops) is $\hat{\mathbf{A}}\in\mathbb{R}^{n\times n}$.
    Besides, $\mathbf{Y} = \{y_1,\dots,y_n\}$ is the label matrix, where $y_v\in\mathbb{R}^{|\mathcal{Y}|}$ is a one-hot vector and $|\mathcal{Y}|$ represents the number of the classes.
    In GU, after receiving unlearning request $\Delta \mathcal{G}=\left\{\Delta \mathcal{V}, \Delta \mathcal{E}, \Delta \mathcal{X}\right\}$ on original model parameterized by $\mathbf{W}$, the goal is to output the predictions of non-unlearning entities (i.e. $\mathcal{V}_U$ ) and adjusted model parameterized by $\mathbf{W}^\star$, both with minimal impact from the unlearning entities.
    The typical unlearning requests include feature-level $\Delta \mathcal{G}=\left\{\varnothing,\varnothing,\Delta \mathcal{X}\right\}$, node-level $\Delta \mathcal{G}=\left\{\Delta \mathcal{V},\varnothing,\varnothing\right\}$, edge-level $\Delta \mathcal{G}=\left\{\varnothing,\Delta \mathcal{E},\varnothing\right\}$ in $\mathcal{V}_L$.

\subsection{Graph Neural Networks}
    Motivated by spectral graph theory and deep neural networks, the concept of graph convolution is initially introduced in~\cite{2013firstgnn}. 
    However, the computational complexity associated with eigenvalue decomposition hinders its deployment. 
    To overcome this challenge, the Graph Convolutional Network (GCN)~\cite{kipf2016gcn} is proposed, which approximates the convolution operator using the first-order approximation of Chebyshev polynomials. 
    GCN propagates node information iteratively to neighboring nodes for label prediction. 
    Building upon this framework, recent studies~\cite{hamilton2017graphsage, velivckovic2017gat, chen2020gcnii} have further optimized the model architectures, achieving remarkable performance improvements.
    Further research advancements on GNNs can be found in recent surveys~\cite{zhou2022gnn_survey2,bessadok2022gnn_survey3,song2022gnn_survey4}.

\begin{figure*}[t]
  \includegraphics[width=\textwidth]{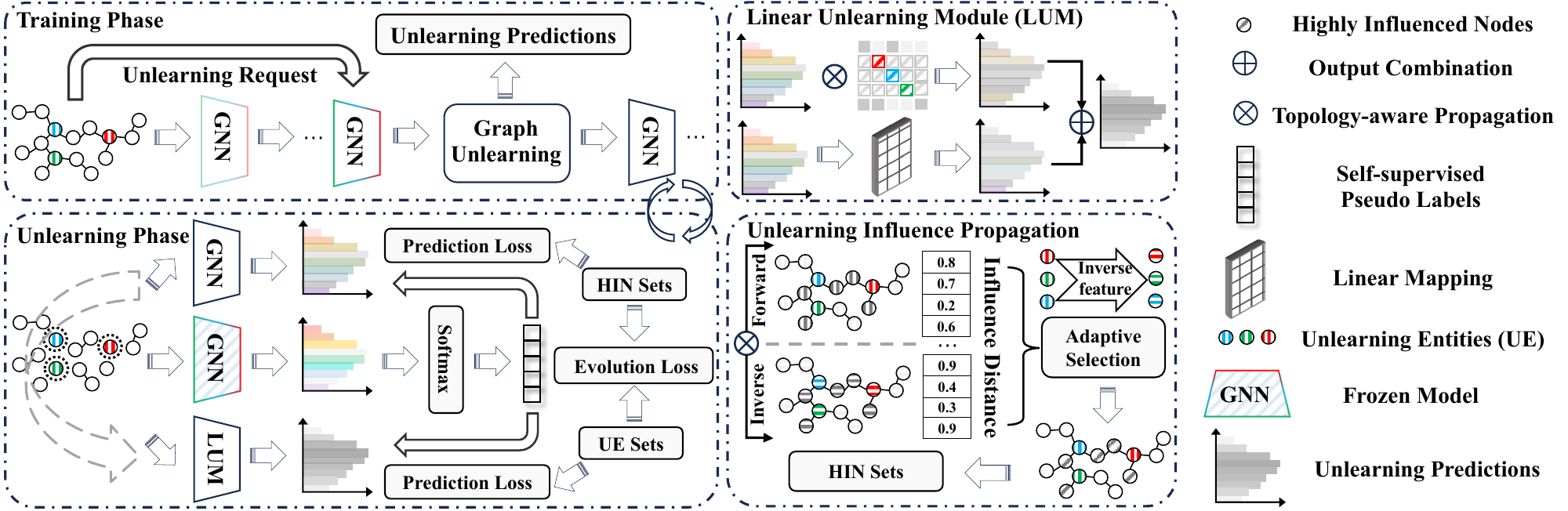}
  \caption{
  Overview of our proposed MEGU.
  Unlearning Prediction represents the prediction of non-unlearning entities.
  }
  \label{fig: framework}
\end{figure*}

\subsection{Graph Unlearning}
\label{sec: graph unlearning}
    In this part, We provide an overview of recent advancements in GU. 
    GraphEraser~\cite{chen2022graph_eraser} attempts to partition the graph into multiple shards to handle unlearning requests within each shard. 
    Building upon this, GUIDE~\cite{wang2023guide} further optimizes the partitioning and shard aggregation strategies. 
    However, their performance depends heavily on partitioning quality and aggregators. 
    GraphEditor~\cite{cong2022grapheditor} and Projector~\cite{cong2023projector} provide closed-form solutions with theoretical guarantees. 
    However, their application is limited due to the linear assumption.
    Approximate-based methods~\cite{chien2022sgc_unlearning, chien2022cgu, wu2023gif} have emerged as efficient solutions.
    However, as highlighted by MEND~\cite{mitchell2021mend}, the lack of consideration for non-unlearning entities may impact their practical performance. 
    Meanwhile, balancing the trade-off during deployment between generalization and performance remains challenging.
    GNNDelete~\cite{cheng2023gnndelete} proposes layer-based unlearning operators to obtain predictions without adjusting the original trained model, but its deployment efficiency decreases with model depth and cannot handle unlearning requests for continue training.

\label{sec: mutual evolution}
\section{Model Framework}
    In this section, we introduce MEGU, which provides a new paradigm for GU by deconstructing the MU targets. 
    To begin with, we provide an overview of the MEGU pipeline and its intuitions. 
    Then, considering the unique challenges posed by GNNs and aiming to achieve graph-based mutual evolution, we introduce adaptive high-influence neighborhood selection and topology-aware unlearning propagation. 
    Building upon these technologies, the predictive module and unlearning module are trained in a topology-guided mutually boosting manner by a well-designed optimization objective.

\subsection{Architecture Overview}
\label{sec: architecture overview}
    As illustrated in Fig.~\ref{fig: framework}, we initialize the predictive module with the original trained model. 
    Throughout the unlearning process, its target is to adjust the original model under unlearning requirements while retaining the reasoning capability. 
    This design preserves the original model's predictive accuracy while efficiently achieving unlearning through an end-to-end learnable mechanism with minimal cost.
    Moreover, the adjusted original model can be further utilized for continued training, offering deployment flexibility.
    As for the unlearning module, its target is to generate predictions for non-unlearning entities based on the predictive module while offering forgetting capacity for model adjustment. 
    This strategy minimizes the computational overhead associated with unlearning. 
    From the mutual evolution perspective, the predictive module relies on the unlearning module's forgetting ability, guiding the modification of the original trained model. 
    Similarly, the unlearning module depends on the predictive module's reasoning capability to generate reliable predictions. 
    Consequently, these two modules mutually optimize each other within the unified MEGU framework.

    For the three downstream unlearning tasks, our processing details are as follows: 
    (1) Feature-level: we treat nodes as unlearning entities while preserving their topology; 
    (2) Node-level: we consider nodes as unlearning entities and remove their related topological connections; 
    (3) Edge-level: we consider connected nodes as unlearning entities but preserve their topology and remove the unlearning edge.

\subsection{Adaptive High-influence Neighborhood Selection}
    Due to the rich interactions in the GNNs, we need to identify the nodes that are highly influenced by unlearning entities.
    This is pivotal in forming an optimization objective that preserves predictive accuracy while reducing unlearning entity impacts.
    Existing methods consider nodes within a fixed neighborhood of unlearning entities as highly influenced nodes (HIN). 
    Unfortunately, they neglect the distinct roles of graph elements in topology-based propagation.
    
    To address this issue, we propose adaptive high-influence neighborhood selection, which leverages the forward and inverse feature propagation based on the original topology to obtain smoothed features from two perspectives.
    Formally, the above process in $l$-layer original GNN can be defined as
    \begin{equation}
    \label{eq: forward_inverse_propagation}
    \begin{aligned}
    &\tilde{\mathbf{X}} =\hat{\mathbf{A}}^{l}\mathbf{X},\;\tilde{\mathbf{X}}' =\hat{\mathbf{A}}^{l}\mathbf{X}',\\
    \mathbf{X}_i'=\mathbf{X}_i,\;\mathbf{X}_j'&=\mathbf{1} - \mathbf{X}_j,\; \forall i\in \mathcal{V}/\Delta \mathcal{V},\; \forall j\in \Delta \mathcal{V},\\
    \end{aligned}
    \end{equation}
    where $\mathbf{1}$ is the 1-vector of size $f$ and $\mathbf{X}'$ is the inverse feature for unlearning entities. 
    Meanwhile, taking into account that the original $l$-layer GNN aggregates information from the $l$-hop neighborhoods, we employ $l$-step feature smoothing by default.
    Notably, in the case of edge unlearning, we treat the two nodes connected by $\Delta \mathcal{E}$ as unlearning entities to perform inverse features.
    Intuitively, when we reverse the features of unlearning entities, it leads to significant changes in the smoothed features of HIN from two topology-based propagation perspectives. 
    To quantify this difference, we introduce the following concept of \textit{influence distance}, which serves as a measure to adaptively select HIN (see Alg.~1). 
    By considering the unique structural properties of different entities, our approach effectively mitigates the bias that arises from treating all nodes within a fixed neighborhood equally.

\begin{definition}{(\textbf{\textit{Influence Distance}}). }
\label{def: influence distance}
    The influence distance $\mathcal{D}_k$ parameterized by node $k$ and forward and inverse feature propagation results $\tilde{\mathbf{X}}$ and $\tilde{\mathbf{X}}'$ is formally defined as
    \begin{equation}
    \label{eq: influence distance}
    \begin{aligned}
    \forall k \in \mathcal{V}/\Delta \mathcal{V},\;\mathcal{D}_k = \mathrm{Dis}(\tilde{\mathbf{X}}_k, \tilde{\mathbf{X}}_k'),\\
    \end{aligned}
    \end{equation}
    where $\tilde{\mathbf{X}}_k$ denotes the $k^{th}$ row of $\tilde{\mathbf{X}}$, $\mathrm{Dis}(\cdot)$ is a function positively relative with the difference, which can be implemented using Euclidean distance, cosine similarity, etc.
\end{definition}

\subsection{Topology-aware Unlearning Propagation}
    To achieve mutual evolution for two individual modules and improve final predictions in graph scenarios, we propose the topology-aware unlearning propagation based on the predictive module and non-unlearning entities $\mathbf{A}^\star,\mathbf{X}^\star$, where we remove the unlearning entities in $\mathbf{A},\mathbf{X}$.
    This strategy considers both the topological structure and the self-supervised information $\mathrm{L}$ from the predictive module, which effectively integrates the predictive and unlearning modules while upholding the homophily assumption to improve predictions.
    Specifically, its foundation lies in the expectation that connected nodes in a graph exhibit similar labels, aligning with the network's inherent homophily or assortative characteristics.
    Thus, we can encourage smoothness over the distribution over labels by another label propagation (i.e. $\mathrm{L}$). 
    Meanwhile, it introduces a novel paradigm for two module interaction in the GU process, which is formally expressed as
    \begin{equation}
    \label{eq: unlearning propagation}
    \begin{aligned}
    &\mathbf{Y}\!\left(\!\hat{\mathbf{Y}}, \mathbf{E}\left(\mathrm{L}\right)\right)\!:=\! \mathbf{Y}_u \!=\!\hat{\mathbf{Y}}_u, \!\mathbf{Y}_v\!=\!\operatorname{G}\!\left(\hat{\mathbf{Y}}_v+\operatorname{G}\left(\mathbf{E}_v\right)\right)\!, \\
    &\mathbf{E}(\mathrm{L})\!:=\! \mathbf{E}^{(0)}_u\! =\!\vec{0}, \mathbf{E}^{(0)}_v \!=\!\mathrm{L}-\hat{\mathbf{Y}}_v, \forall u \in \mathcal{V}_L,\forall v \in \mathcal{V}_U, \\
    &\operatorname{G}\!\left({\mathbf{T}}\right)\!:=\!\mathbf{T}_i^{(l)}\!=\!\alpha{\mathbf{T}}_i^{(0)}\!\!+\!(1\!-\!\alpha)\!\!\!\sum_{j\in\mathcal{N}_i^{(1)}}\!\!\frac{1}{\sqrt{\tilde{d}_i\tilde{d}_j}}{\mathbf{T}}_j^{(l-1)},\\
    \end{aligned}
    \end{equation}
    where $\mathbf{E}$ denotes the error correction matrix.
    Building upon this, we adopt the approximate calculation for the personalized PageRank~\cite{chien2021gprgnn}, where $\mathcal{N}_i^{(1)}$ denotes the one-hop neighbors of node $i$. 
    Meanwhile, we set $\alpha$ according to datasets and backbone-based propagation step $l$ by default to capture structural information.
    The aforementioned process can be regarded as the materialization of the unlearning module leveraging the reasoning capacity of the predictive module to generate reliable predictions.
    As depicted in Fig.~\ref{fig: framework}, this thoughtful technology forms the unlearning module in MEGU, which generates final predictions $\mathbf{Y}^\star$ for non-unlearning entities. 
    It is formally represented as 
    \begin{equation}
    \label{eq: unlearning output}
    \begin{aligned}
    &\;\;\;\;\;\;\;\;\;\;\;\;\;\mathbf{Y}^\star := \mathbf{Y}^\star\left(\hat{\mathbf{Y}}^{\star},\mathbf{E}(\mathrm{\hat{\mathbf{Y}}})\right),\\
    &\hat{\mathbf{P}}=\mathrm{Encoder(\mathbf{A}^\star,\mathbf{X}^\star,\mathbf{W}^\star}),\;\hat{\mathbf{P}}^{\star} = \mathbf{W}_{u}\hat{\mathbf{P}},\\
    &\hat{\mathbf{Y}}=\mathrm{Softmax}\left(\hat{\mathbf{P}}\right),\;\hat{\mathbf{Y}}^{\star} = \mathrm{Softmax}\left(\hat{\mathbf{P}}^{\star}\right),
    \end{aligned}
    \end{equation}
    where $\mathrm{Encoder}(\cdot)$ parameterized by $\mathbf{W}^\star$ is any adjusted original trained model in the predictive module, $\mathbf{W}_u$ is the trainable linear unlearning operator.

\subsection{Optimization Objective}
    Since the unlearning request occurs within the training set, we exclusively utilize self-supervised information during the unlearning process to prevent potential label leakage concerns.
    As illustrated in Fig~\ref{fig: framework}, we freeze the original model at the time of receiving the unlearning request to provide self-supervised information $\tilde{\mathbf{Y}}$, preserving the reasoning and forgetting capacity of the predictive and unlearning module.
    
    Specifically, the predictive module utilizes the cross-entropy (CE) loss based on the output of the frozen model to preserve its reasoning capability. 
    Simultaneously, it leverages Kullback-Leibler divergence (KL) loss and the output of the unlearning module to eliminate the impact of unlearning entities on the original model. 
    Remarkably, benefiting from the initialization of the original model, the predictive module already possesses commendable predictive performance for non-unlearning entities. 
    However, to mitigate the impact of unlearning entities, it is crucial to remove the related knowledge in HIN (KL loss) while maintaining their predictive accuracy (CE loss). 
    Hence, we narrow down the optimization scope of the predictive module from all non-unlearning entities to HIN, which aligns with our dual objectives of unlearning and efficiency improvement.
    \begin{equation}
    \normalsize{
    \label{eq: predictive module loss}
    \begin{aligned}
    \mathcal{L}_{p} = \sum_{u \in \mathrm{HIN}}  \mathcal{L}_{CE}\left(\hat{\mathbf{Y}}_{u}, \tilde{\mathbf{Y}}_{u}\right) + \sum_{v \in \mathrm{HIN}} \mathcal{L}_{KL}(\hat{\mathbf{Y}}_{v}^\star,\hat{\mathbf{Y}}_v).
    \end{aligned}
    }
    \end{equation}

    For the unlearning module, it utilizes the reverse CE loss to enhance its forgetting capability for unlearning entities. 
    Meanwhile, it leverages the KL loss and the output of the predictive module to ensure the predictive performance
    \begin{equation}
    \normalsize{
    \label{eq: unlearning module loss}
    \begin{aligned}
    \mathcal{L}_{u} = \!\!-\!\!\!\!\!\!{\sum_{u \in \Delta \mathcal{V}(\mathcal{X}, \mathcal{E})}}\! \! \!\!\!\!\mathcal{L}_{CE}\!\left(\hat{\mathbf{Y}}_{u}^\star, \tilde{\mathbf{Y}}_{u}\right)\! + \!\!\!\!\!\!\!{\sum_{v \in \Delta \mathcal{V}(\mathcal{X}, \mathcal{E})}}\!\!\!\!\!\!\!\mathcal{L}_{KL}\!(\hat{\mathbf{Y}}_v,\hat{\mathbf{Y}}_v^\star).
    \end{aligned}
    }
    \end{equation}

    Based on Eq.~(\ref{eq: predictive module loss}) and Eq.~(\ref{eq: unlearning module loss}), in the perspective of mutual evolution, we formulate the overall optimization objective in MEGU to achieve $\kappa$-based flexible unlearning 
    \begin{equation}
    \label{eq: megu loss}
    \begin{aligned}
    \mathcal{L} = \mathcal{L}_{p} + \kappa\mathcal{L}_{u}.
    \end{aligned}
    \end{equation}

\begin{algorithm}[t]
\caption{Adaptive HIN Selection}
\label{alg: 1}
	\begin{algorithmic}[1]
		\STATE \textbf{Initialize:} HIN = $\emptyset$, $\omega$ = 0, $\epsilon = 0.1$, $\mu=$True;
		\STATE Execute forward and inverse feature propagation based on the Eq.~(\ref{eq: forward_inverse_propagation}) to obtain $\tilde{\mathbf{X}}, \tilde{\mathbf{X}}'$;
		\STATE Calculate cosine similarity-based influence distance $\mathcal{D}$ according to the Eq.~(\ref{eq: influence distance});
		\WHILE{$\mu$}
                \FOR{node $u$ in $\Delta \mathcal{V}$, node $v$ in $\mathcal{V}/\Delta \mathcal{V}$}
                    \IF{$\mathcal{D}_v \leq \epsilon$ \textnormal{\textbf{and}} $v \in \mathcal{N}^{(l)}_u$}
                        \STATE HIN = HIN $\cup$ $v$;
                    \ENDIF
                \ENDFOR
                \STATE Calculate the maximum $\mathcal{D}_{\textnormal{max}}$ in HIN, $\mu=$False;
                \IF{$\mathcal{D}_{\textnormal{max}} \neq$ $\omega$}
                    \STATE $\omega$ = $\mathcal{D}_{\textnormal{max}}$, $\epsilon = \epsilon +$ 0.1, $\mu=$True;
                \ENDIF   
		\ENDWHILE 
	\end{algorithmic} 
\end{algorithm}

\section{Experiments}
    In this section, we conduct a thorough evaluation of MEGU. 
    We commence by introducing 9 benchmark datasets and baselines.
    Then, we present the methodology used to evaluate the effectiveness of GU.
    Details about the experimental setup can be found in~\cite{MEGU}A.1-A.4.
    In general, we aim to address following questions:
    \textbf{Q1}: Compared to existing GU strategies, can MEGU achieve state-of-the-art performance?
    \textbf{Q2}: If MEGU is effective, where do its reasoning and forgetting capabilities come from?
    \textbf{Q3}: Does MEGU really achieve mutual evolution between the predictive module and unlearning module?
    For more extended experiments and discussions please refer to~\cite{MEGU}A.5-A.6.

\begin{table*}[]
\resizebox{\linewidth}{34mm}{
\setlength{\tabcolsep}{1mm}{

\begin{tabular}{cc|cccccccccccccc}
\midrule[0.3pt]
\multirow{2}{*}{Backbone} & \multirow{2}{*}{Strategy} & \multicolumn{2}{c}{Cora}         & \multicolumn{2}{c}{CiteSeer}     & \multicolumn{2}{c}{PubMed}       & \multicolumn{2}{c}{Photo} & \multicolumn{2}{c}{Computer} & \multicolumn{2}{c}{CS}  & \multicolumn{2}{c}{Physics} \\
                          &                           & F1 Score          & Time         & F1 Score          & Time         & F1 Score          & Time         & F1 Score          & Time         & F1 Score            & Time          & F1 Score          & Time         & F1 Score            & Time           \\ \midrule[0.3pt]
\multirow{8}{*}{GCN}      & Retrain                   & 85.6±0.3          & 14.5         & 75.6±0.2          & 41.0         & 86.5±0.1          & 71.4         & 91.2±0.1          & 39.2         & 83.1±0.2            & 62.7          & 91.4±0.1          & 43.9         & 95.2±0.1            & 169.1          \\
                          & Eraser-LPA                & 42.1±0.0          & 15.4         & 48.0±0.0          & 16.4         & 63.7±0.0          & 33.0         & 45.2±0.0          & 17.7         & 38.2±0.0            & 18.0          & 58.3±0.0          & 24.2         & 65.3±0.0            & 35.5           \\
                          & Eraser-KMeans             & 48.0±0.0          & 14.7         & 39.6±0.0          & 15.7         & 64.4±0.0          & 32.1         & 54.4±0.0          & 17.9         & 40.4±0.0            & 17.9          & 67.0±0.0          & 22.1         & 73.5±0.0            & 33.8           \\
                          & GUIDE-SR                  & 79.2±0.5          & 10.0         & 74.0±0.1          & 11.6         & 85.2±0.0          & 27.9         & 80.5±0.1          & 5.1          & 74.8±0.1            & 9.2           & 85.6±0.1          & 11.2         & 91.6±0.1            & 23.2           \\
                          & GUIDE-Fast                & 79.0±0.2          & 8.9          & 73.6±0.0          & 12.4         & 85.1±0.0          & 28.0         & 80.7±0.0          & 5.3          & 75.9±0.0            & 9.1           & 85.4±0.0          & 11.1         & 91.4±0.0            & 23.0           \\
                          & GIF                       & 83.8±0.3          & 0.3          & 73.9±0.2          & 0.4          & 85.4±0.6          & 0.5          & 89.8±0.3          & 0.3          & 83.2±0.3            & 0.3           & 90.5±0.2          & 0.4          & 93.8±0.1            & \textbf{0.5}   \\
                          & GNNDelete                 & 81.7±0.6          & 1.1          & 72.8±0.4          & 1.1          & 85.0±0.4          & 2.0          & 88.6±0.4          & 1.2          & 83.4±0.2            & 1.3           & 90.7±0.5          & 1.4          & 93.0±0.6            & 1.8            \\
                          & MEGU                      & \textbf{85.2±1.1} & \textbf{0.2} & \textbf{75.8±0.0} & \textbf{0.2} & \textbf{86.9±0.0} & \textbf{0.3} & \textbf{92.2±0.1} & \textbf{0.2} & \textbf{85.6±0.0}   & \textbf{0.2}  & \textbf{92.0±0.0} & \textbf{0.3} & \textbf{95.9±0.0}   & 0.6            \\ \midrule[0.3pt]
\multirow{7}{*}{GAT}      & Retrain                   & 86.3±0.5          & 17.0         & 77.3±0.4          & 43.0         & 86.8±0.2          & 80.9         & 91.8±0.3          & 38.8         & 83.5±0.3            & 63.2          & 91.5±0.2          & 49.9         & 95.4±0.2            & 198.5          \\
                          & Eraser-LPA                & 44.6±0.0          & 23.3         & 48.5±0.0          & 23.2         & 62.5±0.0          & 55.4         & 48.7±0.0          & 28.5         & 40.7±0.0            & 28.4          & 61.3±0.0          & 36.9         & 67.2±0.0            & 63.2           \\
                          & Eraser-KMeans             & 48.3±0.0          & 22.8         & 39.3±0.0          & 23.9         & 64.8±0.0          & 52.0         & 66.0±0.0          & 28.0         & 43.0±0.0            & 28.3          & 70.0±0.0          & 36.1         & 74.0±0.0            & 59.6           \\
                          & GUIDE-SR                  & 76.5±0.5          & 14.6         & 74.1±0.2          & 19.2         & 83.2±0.0          & 38.2         & 81.6±0.1          & 7.6          & 76.5±0.2            & 9.6           & 84.9±0.0          & 13.0         & 89.7±0.1            & 29.7           \\
                          & GUIDE-Fast                & 78.2±0.3          & 15.9         & 74.2±0.2          & 18.2         & 83.4±0.1          & 37.8         & 80.7±0.1          & 6.3          & 76.3±0.2            & 9.1           & 84.8±0.1          & 13.8         & 89.6±0.0            & 28.3           \\
                          & GIF                       & 82.8±0.6          & 0.9          & 73.6±0.2          & 0.8          & 84.5±0.1          & 0.9          & 88.3±0.2          & 0.9          & 82.6±0.3            & 1.1           & 88.3±0.1          & 0.9          & 92.2±0.1            & 1.8            \\
                          & GNNDelete                 & 83.0±0.8          & 1.7          & 73.0±0.5          & 1.5          & 84.7±0.2          & 2.7          & 88.5±0.4          & 1.4          & 82.0±0.3            & 1.6           & 88.5±0.4          & 1.8          & 92.4±0.2            & 2.9            \\
                          & MEGU                      & \textbf{86.4±0.1} & \textbf{0.3} & \textbf{77.8±0.1} & \textbf{0.3} & \textbf{86.2±0.0} & \textbf{0.4} & \textbf{91.5±0.1} & \textbf{0.3} & \textbf{83.8±0.1}   & \textbf{0.5}  & \textbf{91.7±0.1} & \textbf{0.7} & \textbf{95.6±0.1}   & \textbf{1.5}   \\ \midrule[0.3pt]
\end{tabular}
}}
\caption{Transductive performance and training efficiency on the node unlearning.
The best result is {bold}.
}
\label{tab: trans_cmp}
\end{table*}

\begin{table}[t]
\resizebox{\linewidth}{26mm}{
\setlength{\tabcolsep}{2mm}{
\begin{tabular}{cc|cccc}
\midrule[0.3pt]
\multirow{2}{*}{Backbone}    & \multirow{2}{*}{Strategy} & \multicolumn{2}{c}{PPI} & \multicolumn{2}{c}{Flickr} \\
                             &                           & F1 Score    & Time  & F1 Score      & Time   \\ \midrule[0.3pt]
\multirow{4}{*}{GraphSAGE}   & Retrain                   & 56.65±0.20  & 249.2    & 50.64±0.33    & 478.5     \\
                             & GIF                       & 54.23±0.16  & 1.4      & 48.66±0.44    & 1.7       \\
                             & GNNDelete                 & 54.84±0.22  & 9.2      & 48.50±0.56    & 12.6      \\
                             & MEGU                      & \textbf{57.48±0.18} & \textbf{1.3}     & \textbf{50.32±0.36}    & \textbf{1.0}       \\ \midrule[0.3pt]
\multirow{4}{*}{GraphSAINT}  & Retrain                   & 55.32±0.13  & 212.1    & 49.62±0.23    & 402.8     \\
                             & GIF                       & 53.28±0.04  & \textbf{0.5}      & 48.10±0.53    & 1.0       \\
                             & GNNDelete                 & 52.85±0.07  & 5.7      & 47.83±0.42    & 4.8       \\
                             & MEGU                      & \textbf{55.64±0.37}  & {1.0}      & \textbf{49.95±0.54}    & \textbf{0.5}       \\ \midrule[0.3pt]
\multirow{4}{*}{Cluster-GCN} & Retrain                   & 56.37±0.82  & 221.1    & 51.23±0.05    & 425.7     \\
                             & GIF                       & 53.15±1.67  & 0.6      & 48.72±0.69    & 1.0       \\
                             & GNNDelete                 & 54.24±0.98  & 5.7      & 48.55±0.53    & 4.9       \\
                             & MEGU                      & \textbf{57.39±1.02}  & \textbf{0.2}      & \textbf{50.24±0.96}    & \textbf{0.4}       \\ \midrule[0.3pt]
\end{tabular}
}
}
\caption{Inductive performance on the node unlearning.
}
\label{tab: ins_exp}
\end{table}

\begin{table}[h!]
\resizebox{\linewidth}{23.5mm}{
\setlength{\tabcolsep}{1.5mm}{
\begin{tabular}{c|cccc}
\midrule[0.3pt]
\multirow{2}{*}{Strategy} & \multicolumn{2}{c}{PubMed}                & \multicolumn{2}{c}{Flickr} \\
                          & Feature             & Edge                & Feature      & Edge        \\ \midrule[0.3pt]
Retrain                   & 86.85±0.1          & 87.13±0.1          & 48.29±0.2   & 48.14±0.2  \\
Eraser-LPA                & 64.28±0.0          & 64.26±0.0          & 43.51±0.0   & 42.63±0.0  \\
Eraser-Kmeans             & 67.63±0.0          & 65.97±0.1          & 43.18±0.1   & 42.45±0.0  \\
GUIDE-SR                  & 83.73±0.1          & 82.25±0.0          & 46.90±0.0   & 47.02±0.0  \\
GUIDE-Fast                & 83.54±0.0          & 82.32±0.1          & 46.78±0.1   & 46.93±0.1  \\
CGU                       & 79.70±0.1          & 78.31±0.0          & OOT         & OOT           \\
GIF                       & 83.05±0.0          & 82.10±0.1          & 47.09±0.1   & 47.04±0.2  \\
Projector                 & 80.79±0.1          & 81.64±0.1          & 47.06±0.1   & 47.13±0.1  \\
GNNDelete                 & 83.86±0.1          & 82.17±0.1          & 47.12±0.1   & 47.22±0.0           \\
MEGU                      & \textbf{86.95±0.0} & \textbf{86.80±0.0} & \textbf{48.35±0.2}  & \textbf{48.10±0.2}           \\ \midrule[0.3pt]
\end{tabular}
}}
\caption{Predictive performance with SGC backbone.
}
\label{tab: trans_cmp_feature_edge}
\end{table}

\subsection{Experimental Setup}

\textbf{Datasets.}
    We split all datasets following the guidelines of recent GU approaches~\cite{cheng2023gnndelete,wu2023gif}, which randomly split nodes into 80\% for training and 20\% for testing.
    For a comprehensive overview of datasets and baselines, please refer to~\cite{MEGU}A.1.

\textbf{Baselines.}
    We list Retrain and compare MEGU with the following baselines:
    (1) GraphEraser~\cite{chen2022graph_eraser} and GUIDE~\cite{wang2023guide}; 
    (2) CGU~\cite{chien2022cgu} and GIF~\cite{wu2023gif}; 
    (3) Projector~\cite{cong2023projector} and GNNDelete~\cite{cheng2023gnndelete}.
    For details regarding the baselines, please refer to~\cite{MEGU}A.2.
    Unless otherwise stated, we adopt GCN as the backbone and the node unlearning by default to present results.
    Notably, we experiment with multiple backbone GNNs in separate modules to validate the generalizability of MEGU and avoid complex charts, making the results more reader-friendly.
    To alleviate the randomness and ensure a fair comparison, we repeat each experiment 10 times to present unbiased performance.
    We customize the training epochs for each GU strategy to their respective optimal values, ensuring convergence and reporting the average training time (second report).

\textbf{Unlearning Targets.}
    In our experiments, GU requests are categorized as follows:
    (1) Feature-level: We randomly select 10\% of nodes from the training set and mask the full-dimensional features.
    (2) Node-level: We randomly select 10\% of nodes from the training set and remove related edges.
    (3) Edge-level: We randomly select 10\% of edges from the training graph.
    Then, the two nodes connected by the unlearning edges are considered unlearning entities.
    After that, we evaluate the performance of the predictive module using the Micro-F1 score for the semi-supervised node classification, being a harmonic mean of precision and recall, which places greater emphasis on each individual sample.
    As a result, it effectively captures instances of classification errors, making it well-suited for evaluating such unlearning cases.
    Additionally, to verify the forgetting ability of GU strategies, we adopt the Edge Attack. 
    In this strategy, we randomly select two nodes with different labels as targets for adding noisy edges, which are treated as unlearning entities.
    Intuitively, as a method achieves better unlearning, it tends to effectively mitigate the negative impact of noisy edges on predictive performance, thus ensuring robustness.

\begin{figure*}[t]
  \includegraphics[width=\textwidth]{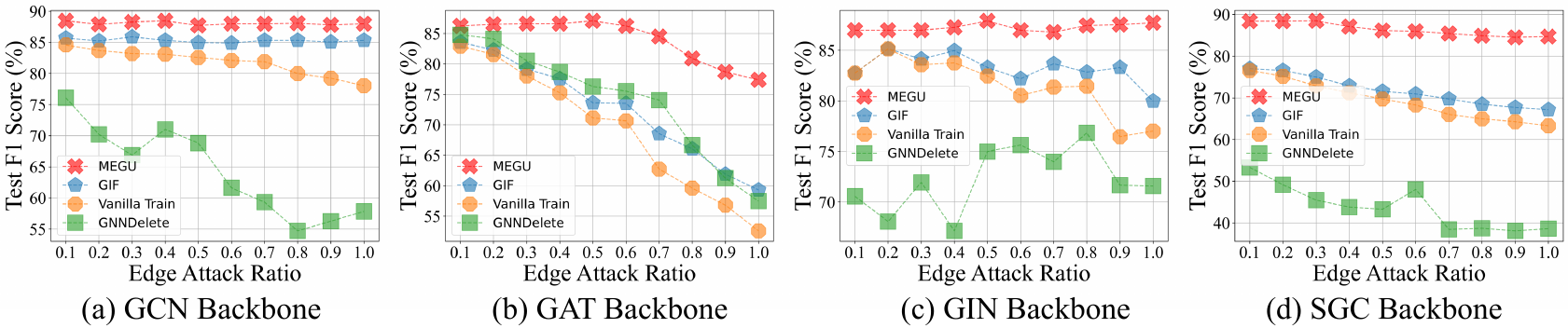}
  \caption{
    Edge Attack performance on Cora.
    The x-axis is the ratio of noisy edges to the existing edges.}
  \label{fig: exp_edge_attack}
\end{figure*}

\begin{table}[t]
\resizebox{\linewidth}{23.5mm}{
\setlength{\tabcolsep}{1.2mm}{
\begin{tabular}{cc|cccc}
\midrule[0.3pt]
\multirow{2}{*}{Model} & \multirow{2}{*}{Component} & \multicolumn{2}{c}{Cora} & \multicolumn{2}{c}{CiteSeer} \\
                       &                            & Feature     & Edge       & Feature       & Edge         \\ \midrule[0.3pt]
\multirow{3}{*}{GCN}   & w/o Ada. HIN                & 87.8±0.6    & 87.3±0.4   & 74.6±0.2      & 76.7±0.3     \\
                       & w/o Topo. UP               & 87.3±0.4    & 87.6±0.3   & 74.5±0.2      & 76.0±0.1     \\
                       & MEGU                       & \textbf{88.5±0.3}    & \textbf{78.5±0.3 }  & \textbf{75.8±0.3 }     & \textbf{77.8±0.1}     \\ \midrule[0.3pt]
\multirow{3}{*}{GAT}   & w/o Ada HIN                & 83.2±0.3    & 84.8±0.5   & 72.8±0.3      & 73.8±0.2     \\
                       & w/o Topo. UP               & 82.8±0.2    & 83.4±0.1   & 73.1±0.1      & 73.6±0.1     \\
                       & MEGU                       & \textbf{84.0±0.3}    & \textbf{85.3±0.2}   & \textbf{74.3±0.2}      & \textbf{74.8±0.1}     \\ \midrule[0.3pt]
\multirow{3}{*}{GIN}   & w/o Ada HIN                & 85.1±0.3    & 84.3±0.4   & 74.4±0.2      & 74.9±0.2     \\
                       & w/o Topo. UP               & 85.7±0.1    & 83.4±0.2   & 74.3±0.1      & 74.7±0.1     \\
                       & MEGU                       & \textbf{86.5±0.2}    & \textbf{75.1±0.1}   & \textbf{75.5±0.2}      & \textbf{75.6±0.1}     \\ \midrule[0.3pt]
\end{tabular}
}}
\caption{Ablation study on three representative backbones.
}
\label{tab: ab}
\end{table}

\subsection{Performance Comparison}
    To answer \textbf{Q1} from the perspective of the predictive module, we report the transductive performance in Table~\ref{tab: trans_cmp}, which validates that MEGU consistently outperforms baselines.
    For instance, on the Cora, MEGU exhibits a remarkable average improvement of 2.4\% over the SOTA approach.
    Notably, the under-performing results of shard-based methods (i.e., GraphEraser and GUIDE) align with their original papers and are likely attributed to the heavy reliance on the partition quality, making them less suitable for scenarios involving substantial element forgetting (10\% unlearning entities). 
    Besides, the results presented in Table~\ref{tab: ins_exp} consistently demonstrate the superior performance of MEGU over all baselines in the inductive setting, underscoring MEGU's remarkable ability to predict unseen nodes.
    Furthermore, we include GU baselines relying on linear model assumptions in Table~\ref{tab: trans_cmp_feature_edge}. 
    Experimental outcomes demonstrate that MEGU outperforms the most competitive methods, achieving average performance gains of 1.7\% and 2.3\% for feature and edge unlearning.
    This observation demonstrates that MEGU can achieve satisfactory performance without relying on a powerful backbone. 
    Its potential for widespread application in linear GNNs is evident, showcasing its generalizability.

    Remarkably, in some cases, MEGU outperforms training GNN from scratch (Retrain). 
    This is because unlearning requests involve the removal of existing graph entities, which could have a negative impact on the Retrain. 
    Fortunately, MEGU's mutual evolution mechanism has the capability to capture such data variations and can mitigate the performance limitations through its optimization framework.

\subsection{Unlearning Capability}
    To answer \textbf{Q1} from the perspective of the unlearning module, we visualize the forgetting capability of various GU strategies under the Edge Attack setting through Fig.~\ref{fig: exp_edge_attack}.
    Intuitively, as the number of noisy edges increases, the accuracy of the unlearning predictions tends to decline. 
    Therefore, for a clear comparison, we introduce vanilla train, a baseline retrained directly on the noisy graph.
    In the context of Edge Attack, we treat noisy edges as unlearning entities. 
    If a GU approach possesses robust unlearning capabilities, it can mitigate the adverse effects caused by noisy edges, thereby ensuring consistent and satisfactory performance.
    From the experimental results, we observe that GNNDelete and GIF do not consistently achieve optimal unlearning performance, whereas MEGU consistently outperforms other baselines in terms of unlearning abilities.
    This advantage is particularly prominent in scenarios where GCN, GIN, and SGC are employed as backbones. 
    Notably, GAT, which heavily relies on edge-based attention mechanisms for information aggregation, is more susceptible to the negative impact of edge attacks, resulting in performance degradation.

\subsection{Training Efficiency} 
\begin{figure*}[t]
  \includegraphics[width=\textwidth]{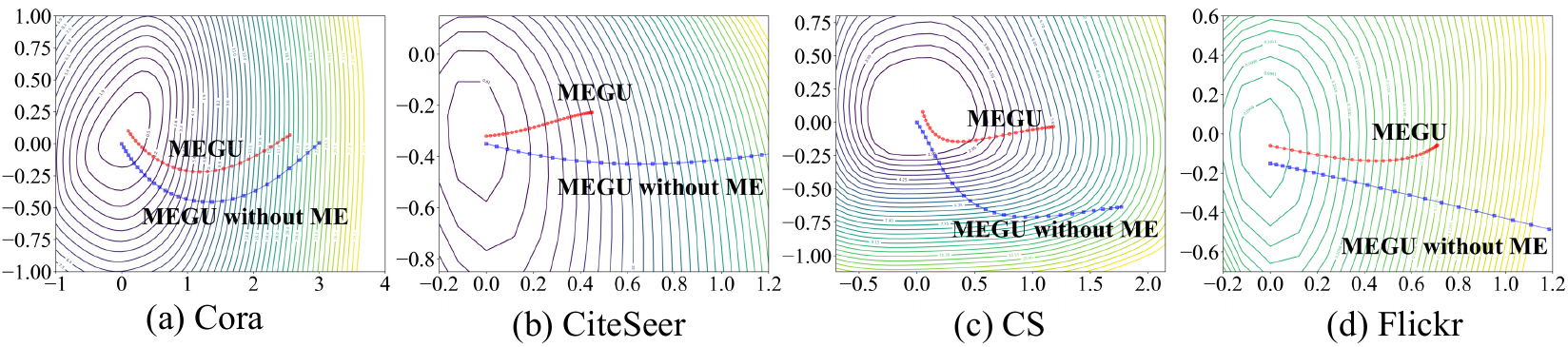}
  \caption{
  The training trajectories of MEGU and its variants without the mutual evolution design on the same loss landscape.}
  \label{fig: exp_trac}
\end{figure*}

    Since GU strategies allow for efficient inference through quick forward computation post-training. 
    Thus, we report the average training time in Table~\ref{tab: trans_cmp} and Table~\ref{tab: ins_exp}.
    In this regard, the pre-training time of the backbone is not included in the report but we incorporate the time required for shard partitioning.
    Notably, to ensure a fair comparison, we customize the training epochs for each GU strategy, guaranteeing model convergence and optimal performance.
    According to the results, our findings are as follows: 
    (1) Shard-based GU methods and retraining GNN from scratch incur significantly long training time;
    (2) Benefiting from the mutual evolution, MEGU achieves model convergence and superior performance within a much shorter time frame (30 - 50 epochs) compared to other strategies. (e.g., GNNDelete requires over 200 epochs)
    This observation is also validated by the experimental results presented in Fig.~\ref{fig: exp_trac}.
    Additionally, Table~\ref{tab: trans_cmp_feature_edge} demonstrates that CGU encounters the OOT (Out of Time) error when dealing with relatively larger-scale graphs (i.e. Flickr), with instances of runtime exceeding 3600 seconds. 
    This arises due to the substantial computational overhead inherent in the process of performing original model corrections based on the gradient Hessian matrix.

\begin{figure}[t]
  \includegraphics[width=\linewidth,scale=1.00]{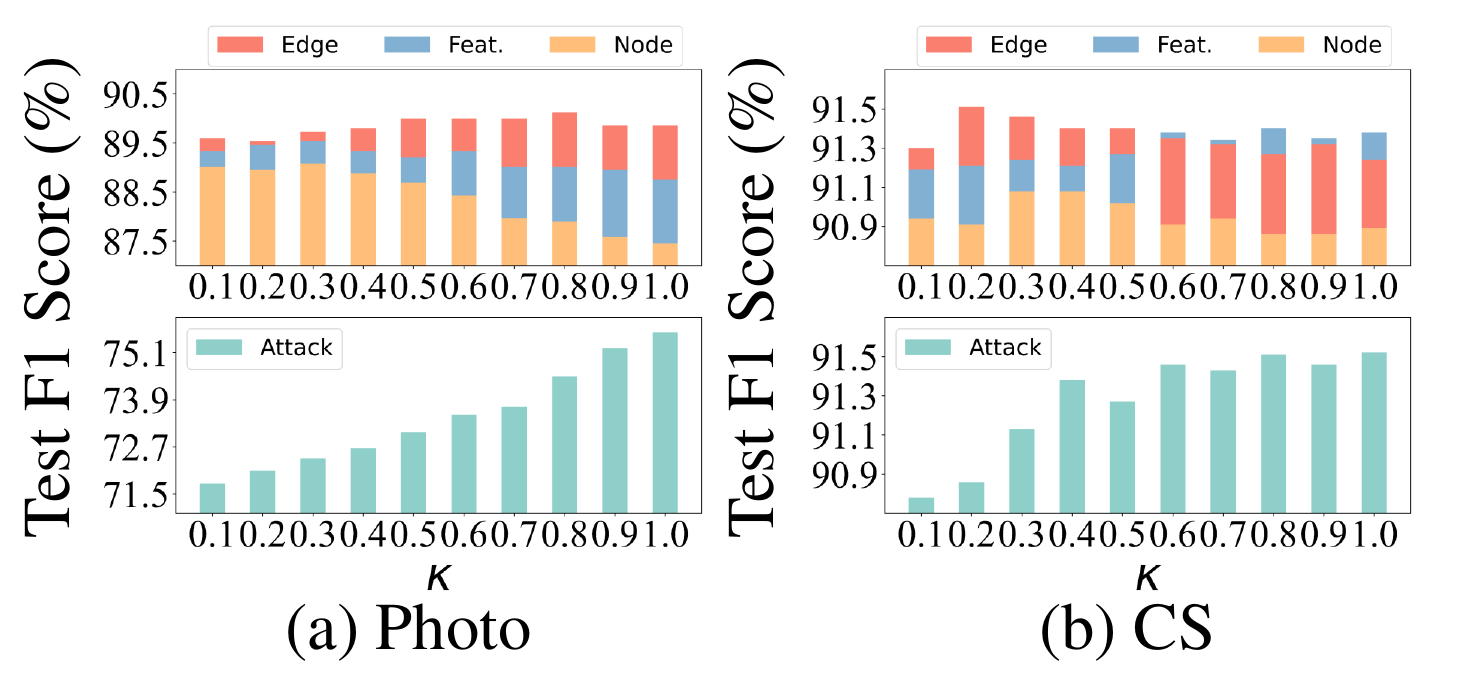}
  \caption{
    Sensitivity analysis on GAT backbone.}
  \label{fig: exp_hyperparameter}
\end{figure}

\subsection{Ablation Study and Sensitivity Analysis}
    To answer \textbf{Q2}, we investigate the contributions of Adaptive HIN Selection (Ada. HIN) and Topology-aware Unlearning Propagation (Topo. UP) in MEGU.
    For Ada. HIN, it constructs a tailored loss function for the predictive module. 
    This ensures the preservation of the predictive module's reasoning ability.
    Results in Table~\ref{tab: ab} show significant performance improvement with Ada. HIN. 
    For instance, in the CiteSeer feature unlearning case using GAT as the backbone, F1 Score increases from 72.8\% to 74.3\%.
    For Topo. UP, it integrates both modules to generate predictions and strengthens their interaction through the mutual evolution loss.
    Experimental results in Table~\ref{tab: ab} highlight Topo. UP's effectiveness in enhancing prediction quality for non-unlearning entities, especially in the GAT backbone. 
    This aligns with our intent to leverage the topology for GU. 
    Topo. UP's propagated features excel in capturing interactions across receptive field sizes, aided by self-supervision.

    In this part, we present the sensitivity analysis in Fig.~\ref{fig: exp_hyperparameter} to further answer \textbf{Q2} from the perspective of the hyperparameter settings (see Eq.~(\ref{eq: megu loss})).
    Based on the experimental results shown in Fig.~\ref{fig: exp_hyperparameter}, we notice that MEGU's predictive performance on non-unlearning entities in feature, node, and edge-level downstream tasks tends to decrease or exhibit unstable fluctuations with increasing $\kappa$. 
    This outcome is attributed to the dilution effect on $\mathcal{L}_p$, which aims to uphold the predictive strength of the predictive module. 
    Conversely, as the emphasis on MEGU's unlearning ability represented by $\mathcal{L}_u$ grows, its performance against Edge Attack progressively improves. 
    These findings offer practical intuition for selecting an appropriate $\kappa$ in real-world scenarios.

\subsection{Mutual Evolution in Graph Unlearning}
    To answer \textbf{Q3}, we visualize the convergence insights into MEGU and its non-mutual evolution variant (MEGU without ME) in the same loss landscape~\cite{li2018visualizing}.
    Fig.~\ref{fig: exp_trac} displays the training trajectories of these two variants, illustrating the convergence states under different training frameworks. 
    In our experimental setup, MEGU without ME implies that the predictive module and unlearning module are independent. 
    Specifically, we remove the additional supervision signal $\hat{\mathbf{P}}$ provided by the predictive module in topology-aware unlearning propagation, as well as the KL Loss that encourages interaction between these two modules in the optimization objective.
    At this time, $\hat{\mathbf{P}}^\star$ in Eq.~(\ref{eq: unlearning propagation}) is generated by the original frozen model.
    Building upon this, we observe that the design of mutual evolution significantly reduces the convergence difficulty and accelerates the convergence speed. 
    This can be validated by the distance between the initial training trajectory point and the global optimal center point, as well as the trajectory itself. 
    Moreover, this observation further elucidates the reason behind MEGU's high training efficiency shown in Table~\ref{tab: trans_cmp} and Table~\ref{tab: ins_exp}, as it achieves optimal performance with a minimal number of training epochs. 
    In a nutshell, the mutual evolution-based GU framework not only mitigates the impact of unlearning entities while improving predictions for non-unlearning entities but also maintains efficient computational performance and flexibility.

\section{Conclusion}
    In this paper, we first address the data removal requirements in graph-based AI applications and provide a new perspective of two crucial modules to analyze the existing GU approaches. 
    Building upon this, we provide reasonable analysis for the essential conditions that GU should satisfy, as illustrated in Table~\ref{tab: gu_methods}. 
    Then, we propose a new framework to achieve effective and general GU via a mutual evolution design. 
    The key insight of our approach lies in leveraging the predictive module's inference capability and the unlearning module's forgetting ability within a unified optimization framework, enabling mutual benefits between the two modules. 
    A promising direction for future GU studies is to explore traceable message-passing mechanisms to further mitigate the impact of unlearning entities and improve predictive performance, allowing both modules to benefit from it.

\section{Acknowledgments}
 This work was partially supported by 
 (I) the National Key Research and Development Program of China 2021YFB3301301, 
 (II) NSFC Grants U2241211, 62072034, and
 (III) High-performance Computing Platform of Peking University.
 Rong-Hua Li is the corresponding author of this paper.

\bibliography{aaai24}

\begin{thebibliography}{31}
\providecommand{\natexlab}[1]{#1}

\bibitem[{Akiba et~al.(2019)Akiba, Sano, Yanase, Ohta, and Koyama}]{akiba2019optuna}
Akiba, T.; Sano, S.; Yanase, T.; Ohta, T.; and Koyama, M. 2019.
\newblock Optuna: A next-generation hyperparameter optimization framework.
\newblock In \emph{Proceedings of the 25th ACM SIGKDD international conference on knowledge discovery \& data mining, KDD}, 2623--2631.

\bibitem[{Bessadok, Mahjoub, and Rekik(2022)}]{bessadok2022gnn_survey3}
Bessadok, A.; Mahjoub, M.~A.; and Rekik, I. 2022.
\newblock Graph neural networks in network neuroscience.
\newblock \emph{IEEE Transactions on Pattern Analysis and Machine Intelligence}, 45(5): 5833--5848.

\bibitem[{Bruna et~al.(2013)Bruna, Zaremba, Szlam, and Lecun}]{2013firstgnn}
Bruna, J.; Zaremba, W.; Szlam, A.; and Lecun, Y. 2013.
\newblock Spectral Networks and Locally Connected Networks on Graphs.
\newblock \emph{Computer Science}.

\bibitem[{Cai et~al.(2021)Cai, Li, Wang, and Ji}]{cai2021link_prediction2}
Cai, L.; Li, J.; Wang, J.; and Ji, S. 2021.
\newblock Line graph neural networks for link prediction.
\newblock \emph{IEEE Transactions on Pattern Analysis and Machine Intelligence}.

\bibitem[{Chen et~al.(2020)Chen, Wei, Huang, Ding, and Li}]{chen2020gcnii}
Chen, M.; Wei, Z.; Huang, Z.; Ding, B.; and Li, Y. 2020.
\newblock Simple and deep graph convolutional networks.
\newblock In \emph{International Conference on Machine Learning, ICML}.

\bibitem[{Chen et~al.(2022)Chen, Zhang, Wang, Backes, Humbert, and Zhang}]{chen2022graph_eraser}
Chen, M.; Zhang, Z.; Wang, T.; Backes, M.; Humbert, M.; and Zhang, Y. 2022.
\newblock Graph unlearning.
\newblock In \emph{Proceedings of the 2022 ACM SIGSAC Conference on Computer and Communications Security, CCS}, 499--513.

\bibitem[{Cheng et~al.(2023)Cheng, Dasoulas, He, Agarwal, and Zitnik}]{cheng2023gnndelete}
Cheng, J.; Dasoulas, G.; He, H.; Agarwal, C.; and Zitnik, M. 2023.
\newblock GNNDelete: A General Strategy for Unlearning in Graph Neural Networks.
\newblock In \emph{The Eleventh International Conference on Learning Representations, ICLR}.

\bibitem[{Chiang et~al.(2019)Chiang, Liu, Si, Li, Bengio, and Hsieh}]{chiang2019cluster-gcn}
Chiang, W.-L.; Liu, X.; Si, S.; Li, Y.; Bengio, S.; and Hsieh, C.-J. 2019.
\newblock Cluster-gcn: An efficient algorithm for training deep and large graph convolutional networks.
\newblock In \emph{Proceedings of the 25th ACM SIGKDD international conference on knowledge discovery \& data mining, KDD}, 257--266.

\bibitem[{Chien, Pan, and Milenkovic(2022)}]{chien2022cgu}
Chien, E.; Pan, C.; and Milenkovic, O. 2022.
\newblock Certified Graph Unlearning.
\newblock In \emph{NeurIPS 2022 Workshop: New Frontiers in Graph Learning}.

\bibitem[{Chien, Pan, and Milenkovic(2023)}]{chien2022sgc_unlearning}
Chien, E.; Pan, C.; and Milenkovic, O. 2023.
\newblock Efficient model updates for approximate unlearning of graph-structured data.
\newblock In \emph{The Eleventh International Conference on Learning Representations, ICLR}.

\bibitem[{Chien et~al.(2021)Chien, Peng, Li, and Milenkovic}]{chien2021gprgnn}
Chien, E.; Peng, J.; Li, P.; and Milenkovic, O. 2021.
\newblock Adaptive Universal Generalized PageRank Graph Neural Network.
\newblock In \emph{International Conference on Learning Representations, ICLR}.

\bibitem[{Cong and Mahdavi(2023{\natexlab{a}})}]{cong2023projector}
Cong, W.; and Mahdavi, M. 2023{\natexlab{a}}.
\newblock Efficiently Forgetting What You Have Learned in Graph Representation Learning via Projection.
\newblock In \emph{International Conference on Artificial Intelligence and Statistics, AISTATS}, 6674--6703. PMLR.

\bibitem[{Cong and Mahdavi(2023{\natexlab{b}})}]{cong2022grapheditor}
Cong, W.; and Mahdavi, M. 2023{\natexlab{b}}.
\newblock GraphEditor: An Efficient Graph Representation Learning and Unlearning Approach.
\newblock \emph{https://openreview.net/forum?id=tyvshLxFUtP}.

\bibitem[{Hamilton, Ying, and Leskovec(2017)}]{hamilton2017graphsage}
Hamilton, W.; Ying, Z.; and Leskovec, J. 2017.
\newblock Inductive representation learning on large graphs.
\newblock \emph{Advances in Neural Information Processing Systems, NeurIPS}.

\bibitem[{Kipf and Welling(2017)}]{kipf2016gcn}
Kipf, T.~N.; and Welling, M. 2017.
\newblock Semi-supervised classification with graph convolutional networks.
\newblock In \emph{International Conference on Learning Representations, ICLR}.

\bibitem[{Li et~al.(2018)Li, Xu, Taylor, Studer, and Goldstein}]{li2018visualizing}
Li, H.; Xu, Z.; Taylor, G.; Studer, C.; and Goldstein, T. 2018.
\newblock Visualizing the loss landscape of neural nets.
\newblock \emph{Advances in neural information processing systems, NeurIPS}, 31.

\bibitem[{Li et~al.(2023)Li, Zhao, Wu, Zhang, Li, and Wang}]{MEGU}
Li, X.; Zhao, Y.; Wu, Z.; Zhang, W.; Li, R.-H.; and Wang, G. 2023.
\newblock MEGU Technical Report.
\newblock In \emph{https://github.com/xkLi-Allen/MEGU}.

\bibitem[{Mitchell et~al.(2022)Mitchell, Lin, Bosselut, Finn, and Manning}]{mitchell2021mend}
Mitchell, E.; Lin, C.; Bosselut, A.; Finn, C.; and Manning, C.~D. 2022.
\newblock Fast model editing at scale.
\newblock In \emph{The Tenth International Conference on Learning Representations, ICLR}.

\bibitem[{Shchur et~al.(2018)Shchur, Mumme, Bojchevski, and G{\"u}nnemann}]{shchur2018amazon_datasets}
Shchur, O.; Mumme, M.; Bojchevski, A.; and G{\"u}nnemann, S. 2018.
\newblock Pitfalls of graph neural network evaluation.
\newblock \emph{arXiv preprint arXiv:1811.05868}.

\bibitem[{Song et~al.(2022)Song, Yang, Xu, and King}]{song2022gnn_survey4}
Song, Z.; Yang, X.; Xu, Z.; and King, I. 2022.
\newblock Graph-based semi-supervised learning: A comprehensive review.
\newblock \emph{IEEE Transactions on Neural Networks and Learning Systems}.

\bibitem[{Tan et~al.(2023)Tan, Zhang, Liu, Zha, Li, Chen, Choi, and Hu}]{tan2023link_prediction4}
Tan, Q.; Zhang, X.; Liu, N.; Zha, D.; Li, L.; Chen, R.; Choi, S.-H.; and Hu, X. 2023.
\newblock Bring your own view: Graph neural networks for link prediction with personalized subgraph selection.
\newblock In \emph{Proceedings of the Sixteenth ACM International Conference on Web Search and Data Mining, WSDM}, 625--633.

\bibitem[{Veli{\v{c}}kovi{\'c} et~al.(2018)Veli{\v{c}}kovi{\'c}, Cucurull, Casanova, Romero, Lio, and Bengio}]{velivckovic2017gat}
Veli{\v{c}}kovi{\'c}, P.; Cucurull, G.; Casanova, A.; Romero, A.; Lio, P.; and Bengio, Y. 2018.
\newblock Graph attention networks.
\newblock In \emph{International Conference on Learning Representations, ICLR}.

\bibitem[{Wang, Huai, and Wang(2023)}]{wang2023guide}
Wang, C.-L.; Huai, M.; and Wang, D. 2023.
\newblock Inductive Graph Unlearning.
\newblock \emph{arXiv preprint arXiv:2304.03093}.

\bibitem[{Wu et~al.(2019)Wu, Souza, Zhang, Fifty, Yu, and Weinberger}]{wu2019sgc}
Wu, F.; Souza, A.; Zhang, T.; Fifty, C.; Yu, T.; and Weinberger, K. 2019.
\newblock Simplifying graph convolutional networks.
\newblock In \emph{International conference on machine learning, ICML}.

\bibitem[{Wu et~al.(2023)Wu, Yang, Qian, Sui, Wang, and He}]{wu2023gif}
Wu, J.; Yang, Y.; Qian, Y.; Sui, Y.; Wang, X.; and He, X. 2023.
\newblock GIF: A General Graph Unlearning Strategy via Influence Function.
\newblock In \emph{Proceedings of the ACM Web Conference, WWW}, 651--661.

\bibitem[{Xu et~al.(2019)Xu, Hu, Leskovec, and Jegelka}]{xu2018gin}
Xu, K.; Hu, W.; Leskovec, J.; and Jegelka, S. 2019.
\newblock How Powerful are Graph Neural Networks?
\newblock In \emph{International Conference on Learning Representations, ICLR}.

\bibitem[{Yang et~al.(2022)Yang, Shen, Li, Qi, Zhang, and Yin}]{yang2022graph_classification3}
Yang, M.; Shen, Y.; Li, R.; Qi, H.; Zhang, Q.; and Yin, B. 2022.
\newblock A new perspective on the effects of spectrum in graph neural networks.
\newblock In \emph{International Conference on Machine Learning, ICML}, 25261--25279. PMLR.

\bibitem[{Yang, Cohen, and Salakhutdinov(2016)}]{Yang16cora}
Yang, Z.; Cohen, W.~W.; and Salakhutdinov, R. 2016.
\newblock Revisiting Semi-Supervised Learning with Graph Embeddings.
\newblock In \emph{Proceedings of the 33rd International Conference on International Conference on Machine Learning, ICML}, 40–48.

\bibitem[{Zeng et~al.(2020)Zeng, Zhou, Srivastava, Kannan, and Prasanna}]{zeng2019graphsaint}
Zeng, H.; Zhou, H.; Srivastava, A.; Kannan, R.; and Prasanna, V. 2020.
\newblock Graphsaint: Graph sampling based inductive learning method.
\newblock In \emph{International conference on learning representations, ICLR}.

\bibitem[{Zhang et~al.(2022)Zhang, Yin, Sheng, Li, Ouyang, Li, Tao, Yang, and Cui}]{gamlp}
Zhang, W.; Yin, Z.; Sheng, Z.; Li, Y.; Ouyang, W.; Li, X.; Tao, Y.; Yang, Z.; and Cui, B. 2022.
\newblock Graph Attention Multi-Layer Perceptron.
\newblock \emph{Proceedings of the 28th ACM SIGKDD Conference on Knowledge Discovery and Data Mining, KDD}.

\bibitem[{Zhou et~al.(2022)Zhou, Zheng, Huang, Hao, Li, and Zhao}]{zhou2022gnn_survey2}
Zhou, Y.; Zheng, H.; Huang, X.; Hao, S.; Li, D.; and Zhao, J. 2022.
\newblock Graph Neural Networks: Taxonomy, Advances, and Trends.
\newblock \emph{ACM Transactions on Intelligent Systems and Technology (TIST)}, 13(1): 1--54.

\end{thebibliography}

\clearpage
\appendix

\section{Outline}
The appendix is organized as follows:
\begin{description}
    \item[A.1] Dataset Description.
    \item[A.2] Compared Baselines.
    \item[A.3] Hyperparameter settings.
    \item[A.4] Experiment Environment.
    \item[A.5] Sparsity Challenge in Feature Unlearning.
    \item[A.6] Unlearning Challenges at Different Scales.
\end{description}

\subsection{A.1 Dataset Description}

\begin{table*}[t]
    \caption{The statistics of the experimental datasets.
    }
    \label{tab: datasets}
    \resizebox{\linewidth}{25mm}{
    \setlength{\tabcolsep}{1.5mm}{
    \begin{tabular}{cccccccc}
    \midrule[0.3pt]
    Dataset          & \#Nodes & \#Features & \#Edges    & \#Classes & \#Feat./Node/Edge Unlearn    & Task type    & Description         \\ \midrule[0.3pt]
    Cora             & 2,708   & 1,433      & 5,429      & 7         & 216/216/802                  & Transductive & citation network    \\
    CiteSeer         & 3,327   & 3,703      & 4,732      & 6         & 266/266/736                  & Transductive & citation network    \\
    PubMed           & 19,717  & 500        & 44,338     & 3         & 1,577/1,577/5,426            & Transductive & citation network    \\ \midrule[0.3pt]
    Amazon Photo     & 7,487   & 745        & 119,043    & 8         & 612/612/5,889                & Transductive & co-purchase graph \\ 
    Amazon Computers  & 13,381  & 767        & 245,778    & 10        & 1100/1100/10651              & Transductive & co-purchase graph \\ \midrule[0.3pt]
    Coauthor CS      & 18,333  & 6,805      & 81,894     & 15        & 1,466/1,466/9,081            & Transductive & co-authorship graph \\
    Coauthor Physics & 34,493  & 8,415      & 247,962    & 5         & 2,759/2,759/21,712           & Transductive & co-authorship graph \\ \midrule[0.3pt]
    PPI              & 56,944  & 50         & 818,716    & 121       & 4,555/4,555/39,993           & Inductive    & protein interactions network     \\
    Flickr           & 89,250  & 500        & 899,756    & 7         & 7,140/7,140/47,449           & Inductive    & image network       \\ \midrule[0.3pt]
    \end{tabular}
    }}
\end{table*}

    The 9 statistics of graph benchmark datasets and feature, node, edge-level unlearning requests are shown in Table.\ref{tab: datasets}.
    Moreover, the description of all datasets are listed below:

    \textbf{Cora}, \textbf{CiteSeer}, and \textbf{PubMed}~\cite{Yang16cora} are three citation network datasets representing undirected graphs, where nodes represent papers and edges represent citation relationships between papers. 
    The node features are word vectors, where each element is a binary variable (0 or 1) indicating the presence or absence of each word in the paper. 

    \textbf{Coauthor CS} and \textbf{Coauthor Physics}~\cite{shchur2018amazon_datasets} are co-authorship graphs based on the Microsoft Academic Graph. 
    Here, nodes are authors, that are connected by an edge if they co-authored a paper; node features represent paper keywords for each author’s papers, and class labels indicate the most active fields of study for each author.

    \textbf{Amazon Photo} and \textbf{Amazon Computers}~\cite{shchur2018amazon_datasets} are segments of the Amazon co-purchase graph. Nodes represent goods and edges represent that two goods are frequently bought together. Given product reviews as bag-of-words node features, the task is to map goods to their respective product category.

    \textbf{PPI}~\cite{zeng2019graphsaint} stands for Protein-Protein Interaction (PPI) network, where nodes represent protein.  
    If two proteins participate in a life process or perform a certain function together, it is regarded as an interaction between these two proteins. 
    Complex interactions between multiple proteins can be described by PPI networks.

    \textbf{Flickr}~\cite{zeng2019graphsaint} dataset originates from the SNAP, they collect Flickr data and generate an undirected graph. 
    Nodes represent images, and edges connect images with common properties like geographic location, gallery, or shared comments. 
    Node features are 500-dimensional bag-of-words representations extracted from the images. 
    The labels are manually merged from the 81 tags into 7 classes.

\subsection{A.2 Compared Baselines}
    \textbf{Backbone GNNs.} 
    To evaluate the effectiveness of various GU strategies, we have selected commonly used GNNs as the backbone models to simulate scenarios where unlearning requests are received during training. 
    The chosen models encompass GCN~\cite{kipf2016gcn}, GAT~\cite{velivckovic2017gat}, GraphSage~\cite{hamilton2017graphsage}, GIN~\cite{xu2018gin}, SGC~\cite{wu2019sgc}, Cluster-GCN~\cite{chiang2019cluster-gcn}, and GraphSAINT~\cite{zeng2019graphsaint}. 
    These models represent successful recent designs in graph learning, widely applicable in both transductive and inductive settings. 
    Furthermore, various backbone GNNs can be employed to assess the generalization capabilities of diverse GU approaches.
    The salient characteristics of all baseline models are outlined below:
    
    \textbf{GCN}~\cite{kipf2016gcn} introduces a novel approach to graph-structured data that uses an efficient layer-wise propagation rule that is based on a first-order approximation of spectral convolutions on graphs.

    \textbf{GAT}~\cite{velivckovic2017gat} utilizes a graph attention layer to assign varying importance to different nodes within a neighborhood, thus better-representing graph information.

    \textbf{GIN}~\cite{xu2018gin} develops a simple graph learning architecture with MLP that is as powerful as the Weisfeiler-Lehman graph isomorphism test.  

    \textbf{SGC}~\cite{wu2019sgc} simplifies GCN by removing nonlinearities and collapsing weight matrices between consecutive layers, bringing higher running efficiency.

    \textbf{GraphSage}~\cite{hamilton2017graphsage} is an inductive framework that leverages neighbor node attribute information to efficiently generate representations.

    \textbf{Cluster-GCN}~\cite{chiang2019cluster-gcn} is a novel GNN designed for training with Stochastic Gradient Descent (SGD) by leveraging the graph clustering structure. 

    \textbf{GraphSAINT}~\cite{zeng2019graphsaint} is a novel inductive learning method that enhances training efficiency and accuracy through graph sampling.

     \textbf{Graph Unlearning strategies.} 
     In our experimental study, we delineate the characteristics and provide descriptions of GU strategies that have been proposed in recent years:

    \textbf{GraphEraser}~\cite{chen2022graph_eraser} propose a novel machine unlearning framework tailored to graph data. Its contributions include two novel graph partition algorithms and a learning-based aggregation method.

    \textbf{GUIDE}~\cite{wang2023guide} improves GraphEraser by the graph partitioning with fairness and balance, efficient subgraph repair, and similarity-based aggregation.

    \textbf{CGU}~\cite{chien2022cgu} presents the underlying analysis of certified GU using SGC and their generalized PageRank (GPR) extensions as examples. 

    \textbf{GIF}~\cite{wu2023gif} incorporates an additional loss term for influenced neighbors, considering structural dependencies, and provides a closed-form solution for better understanding the unlearning mechanism.

    \textbf{Projector}~\cite{cong2023projector} achieves unlearning by projecting the weights of the pre-trained linear model onto a subspace that is unrelated to the unlearning entities.

    \textbf{GNNDelete}~\cite{cheng2023gnndelete} is a novel model-agnostic layer-wise operator designed to optimize topology influence in the graph unlearning requests.

\begin{table}[t]
\caption{Detailed hyperparameter setting on all datasets.}
\centering
\label{tab: exp_unlearn_hyperparameters}
\resizebox{0.9\linewidth}{25mm}{
\setlength{\tabcolsep}{1.2mm}{
\begin{tabular}{ccccc}
\midrule[0.3pt]
Dataset          & unlearning rate & $\kappa$   & $\alpha_1$ & $\alpha_2$  \\ \midrule[0.3pt]
Cora             & 0.05            & 0.01       & 0.8               & 0.5   \\ 
CiteSeer         & 0.09            & 0.01       & 0.24              & 0.12  \\ 
PubMed           & 0.04            & 0.09       & 0.18              & 0.12  \\ \midrule[0.3pt]
Amazon Photo     & 0.065           & 0.06       & 0.94              & 0.2   \\ 
Amazon Computers & 0.001           & 0.01       & 0.05              & 0.05  \\ \midrule[0.3pt]
Coauthor CS      & 0.007           & 0.01       & 0.03              & 0.13  \\
Coauthor Physics & 0.04            & 0.1        & 0.02              & 0.27   \\ \midrule[0.3pt]
PPI              & 0.03            & 0.08       & -                 & -  \\
Flickr           & 0.001           & 0.01       & 0.05              & 0.05  \\ \midrule[0.3pt]
\end{tabular}}}
\end{table}

\begin{figure*}[t]
    \centering
    \includegraphics[width=\textwidth]{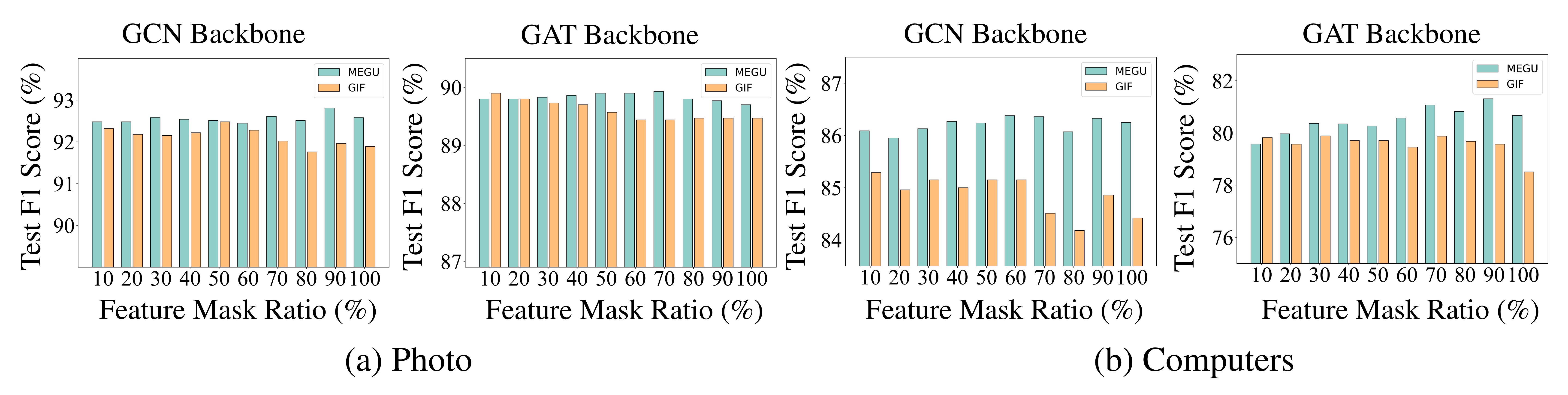}
    \caption{Performance of different feature mask ratios on Photo and Computers with GCN and GAT backbone.}
    \label{fig: exp_supple_feature_mask_bar}
\end{figure*}

\begin{figure*}[t]
    \centering
    \includegraphics[width=\textwidth]{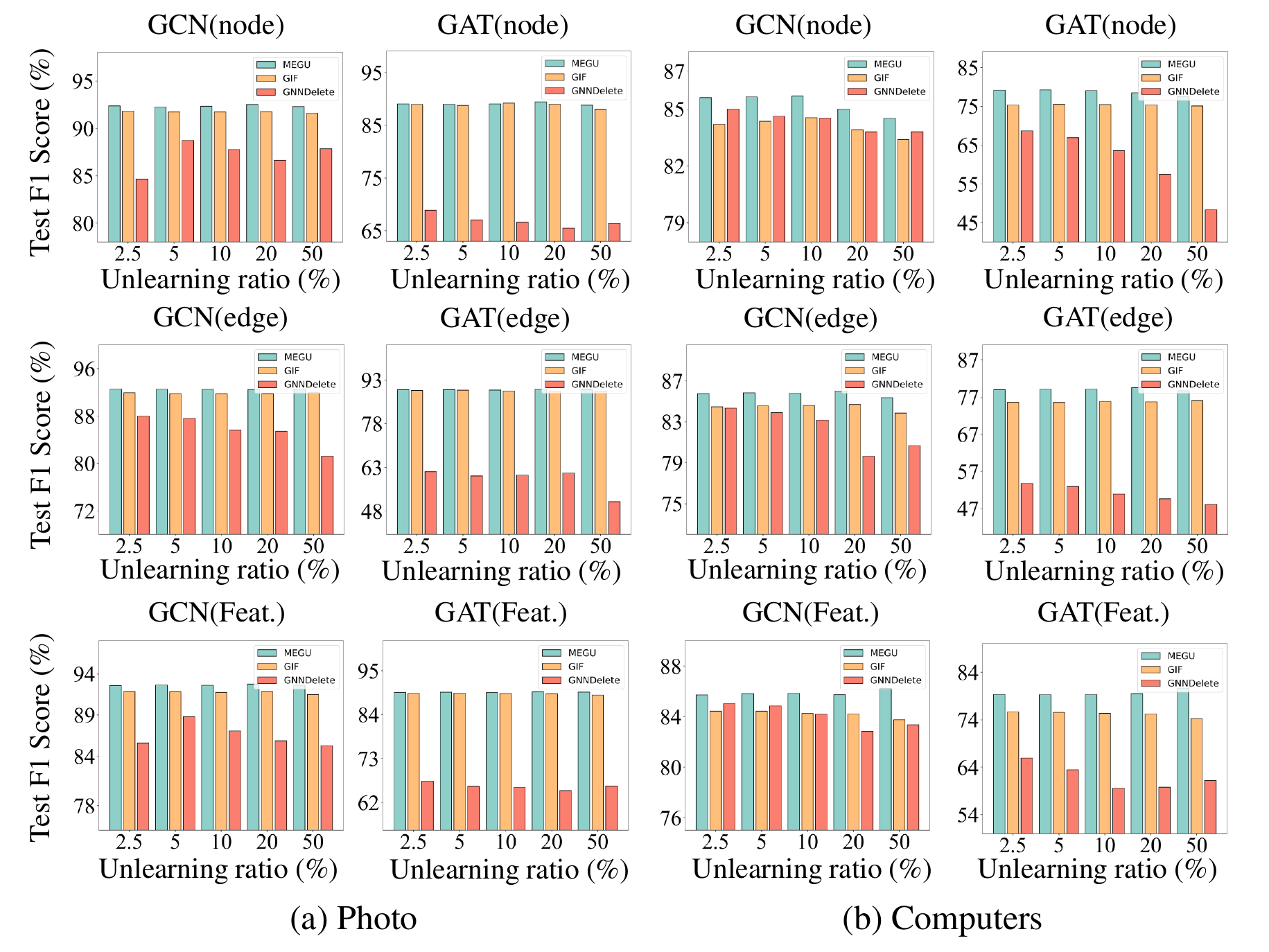}
    \caption{Performance of multiple unlearning entities and different unlearning ratios on Photo and Computer.}
    \label{fig: exp_supple_bar}
\end{figure*}

\subsection{A.3 Hyperparameter settings}
    The hyperparameters in the backbones and GU approaches are set according to the original paper if available.
    Otherwise, we perform an automatic hyperparameter search via the Optuna~\cite{akiba2019optuna}.
    Specifically, we explore the optimal shards within the ranges of 20 to 100.
    The weight coefficients of the loss function and other hyperparameter is get by means of an interval search from $\{0, 1\}$ or the interval suggested in the original paper. 
    For our proposed MEGU, the coefficient of personalized PageRank in the context of the topology-aware unlearning propagation process ($\alpha$) and loss function ($\kappa$) are explored within the ranges of 0 to 1.
    More details can be referred to Eq.~(\ref{eq: unlearning propagation}) and Eq.~(\ref{eq: megu loss}).

    To help reproduce the experimental results, we provide the hyperparameter settings in Table \ref{tab: exp_unlearn_hyperparameters}, where $\alpha_1$ and $\alpha_2$ correspond to the $\mathbf{E}_v$ and $\hat{\mathbf{Y}}_v+\mathbf{E}_v$ propagation coefficients in Eq.~(\ref{eq: unlearning propagation}). 
    The hyperparameters presented in this table are applicable to all backbones mentioned in this paper. 
    Since the PPI dataset is multi-label classification task, in order to avoid propagating high bias on the graph due to multi-label classification, we did not use the Topo. UP module when processing this dataset, therefore the PPI dataset does not have the corresponding $\alpha_1$ and $\alpha_2$. 
    In addition, we use SGD as the optimizer and set the number of epochs to 100. Specific experimental strategies and examples can be found in https://github.com/xkLi-Allen/MEGU.

\subsection{A.4 Experiment Environment}
    Experiments are conducted with Intel(R) Xeon(R) CPU E5-2686 v4 @ 2.30GHz, and a single NVIDIA GeForce RTX 3090 with 24GB GPU memory. 
    The operating system of the machine is Ubuntu 20.04.5. 
    As for software versions, we use Python 3.8.10, Pytorch 1.13.0, and CUDA 11.7.0.

\subsection{A.5 Sparsity Challenge in Feature Unlearning}
    In the context of feature sparsity, we posit that the feature representation of labeled nodes is partially incomplete. 
    In the context of feature unlearning, these labeled nodes correspond to unlearning entities are afflicted by feature-related noise, while the incompleteness of feature representation aligns with the objective of feature unlearning.
    This necessitates mitigating the impact of unlearning features on other entities within the graph learning paradigm. 
    In the main text, we consider masking all dimensions of features for the unlearning nodes to evaluate the feature unlearning performance of different GU strategies. 
    However, such a choice may not encompass the entirety of the feature unlearning.
    To further elucidate the superior performance of MEGU in the realm of feature unlearning, we expand the experimental scope of the feature unlearning. 
    Multiple experiments are conducted using feature masking ratios ranging from 0.1 to 1, and the obtained results are juxtaposed with those of the most competitive GIF, as illustrated in Fig.~\ref{fig: exp_supple_feature_mask_bar}.
    
    Building upon this, our findings can be summarized as follows:
    (1)
    The feature mask ratio exerts a substantial influence on GU performance. 
    As the feature mask ratio increases, a diminishing and unstable performance trend becomes evident across various GU strategies, particularly pronounced in the context of GIF. 
    This phenomenon is attributed to the heightened feature mask ratio intensifying the unlearning cost within the GU framework, thereby presenting a complex trade-off between predictive accuracy for non-unlearning entities and the efficacy of forgetting unlearning entities.
    (2) Notable advantages of MEGU.
    As depicted in Fig. \ref{fig: exp_supple_feature_mask_bar}, a clear trend emerges wherein the performance of the unlearned model derived from GIF demonstrates a decline with increasing feature mask ratio. 
    In contrast, the performance of the unlearned model obtained through MEGU maintains a consistently superior level. 
    This resilience is attributed to MEGU's incorporation of a mutual evolution mechanism, which orchestrates a harmonious equilibrium between the predictive and unlearning modules.

\subsection{A.6 Unlearning Challenges at Different Scales}
    In our experimental setup outlined in the main text, we adopt a default configuration wherein 10\% of the graph elements are chosen as unlearning entities. 
    In order to comprehensively evaluate the efficacy of MEGU across varying unlearning scales, we present additional experimental results in Fig. \ref{fig: exp_supple_bar}. 
    According to our experiments, we observe that feature unlearning demonstrates with much less impact by the scales of unleanring tasks compared to node and edge unlearning. 
    This discrepancy arises from the nuanced process associated with node unlearning, wherein the edges directly connected to the unlearning node are expunged, thereby disrupting the topology and inducing performance deterioration. 
    In the edge unlearning scenarios, we remove the unlearning edges and treat the nodes connected through those edges as the entities, entailing a heightened unlearning cost.
    In summation, feature unlearning induces a comparatively milder impact on predictive performance for non-unlearning entities when compared to the other unlearning scenarios.
    
    MEGU excels in accommodating diverse unlearning requests. 
    Most evidently, as the unlearning ratio increases, an inevitable decline in the performance of the unlearned model becomes apparent across the three GU methodologies. 
    This decline is attributed to the increased ratio of forgetten data, which thereby magnifies the negative impact on the model predictive performance, and consequently leads to a gradual erosion of the performances. 
    In this context, it becomes evident that MEGU outperforms GNNDelete and GIF under identical unlearning conditions. 
    These empirical findings and analyses underscore MEGU's capability to address unlearning tasks at the node, edge, and feature levels.

\end{document}